\documentclass[10pt,twocolumn,letterpaper]{article}

\usepackage{3dv}
\usepackage{times}
\usepackage{epsfig}
\usepackage{graphicx}
\usepackage{amsmath}
\usepackage{amssymb}
\usepackage{appendix}

\usepackage[pagebackref=true,breaklinks=true,letterpaper=true,colorlinks,bookmarks=false]{hyperref}

\threedvfinalcopy 

\def\threedvPaperID{100} 

\ifthreedvfinal\fi
\begin{document}

\title{SAFA: Structure Aware Face Animation}

\author{Qiulin Wang\\
JD Technology\\
{\tt\small wangqiulin5@jd.com}
\and
Lu Zhang\\
JD Technology\\
{\tt\small zhanglu32@jd.com}
\and
Bo Li\\
JD Technology\\
{\tt\small prclibo@gmail.com}
}

\maketitle
\thispagestyle{empty}

\begin{abstract}
Recent success of generative adversarial networks (GAN) has made great progress on the face animation task. However, the complex scene structure of a face image still makes it a challenge to generate videos with face poses significantly deviating from the source image. On one hand, without knowing the facial geometric structure, generated face images might be improperly distorted. On the other hand, some area of the generated image might be occluded in the source image, which makes it difficult for GAN to generate realistic appearance. To address these problems, we propose a structure aware face animation (SAFA) method which constructs specific geometric structures to model different components of a face image. Following the well recognized motion based face animation technique, we use a 3D morphable model (3DMM) to model the face, multiple affine transforms to model the other foreground components like hair and beard, and an identity transform to model the background. The 3DMM geometric embedding not only helps generate realistic structure for the driving scene, but also contributes to better perception of occluded area in the generated image. Besides, we further propose to exploit the widely studied inpainting technique to faithfully recover the occluded image area. Both quantitative and qualitative experiment results have shown the superiority of our method. Code is available at \url{https://github.com/Qiulin-W/SAFA}.
\end{abstract}

\section{Introduction}
Face animation refers to the task to generate a face video from a source face image to reenact the face in a given driving video, possibly of different identity from the source. The source image provides the appearance and identity information and the driving video determines face poses and translational motions. Generally, a generated face video with high quality should meet multiple requirements including but not limited to the following:

(1) \textbf{pose-preserving}: The generated face video need accurately follow the specific pose, expression and motion of the driving video. 

(2) \textbf{identity-preserving}: The generated face video should preserve the identity information of the source face.

(3) \textbf{realistic}: The generated face images should possess realistic face shape, texture and visual sharpness.

(4) \textbf{occlusion aware}: When an area in the generated image is occluded in the source image, the model should be aware of the occlusion and recover the missing part. Such occlusion might occur on the foreground face when a head turns or on the background when a head undergoes translational move.

With the rapid development of Generative Adversarial Networks (GANs)~\cite{goodfellow2014generative}, recent deep learning based methods~\cite{wiles2018x2face, kim2018deep, wang2019few, zakharov2019few, zakharov2020fast, siarohin2020first, MarioNETte:AAAI2020, doukas2020headgan, wang2020one} have shown promising results in generating face animation videos. However, it is still challenging to meet all the four requirements stated above especially when large face pose variation, translational/expressional motion and occlusion occur. 

In this paper, we resolve the above challenges by integrating the scene structure knowledge in face animation procedure. The scene structure can be decomposed into foreground and background. The foreground refers to the head which can be further decomposed into face and other components like hair, beard and clothes. The face region is conventionally considered as a rigid body in computer vision while others are relatively soft. The background is static in the typical scenario of face animation. It is recognized to be difficult to learn this complex structure elaborately in the end-to-end manner and specific priors is essential for high quality generation. \cite{siarohin2020first} proposes a First Order Motion Model (FOMM) to model foreground and background motion by affine motion and static motion, respectively. However, when face pose or expression changes significantly between source and driving images, 2D affine motion is insufficient to model such transforms. 

To address this problem, we integrate the state-of-the-art 3D morphable model (3DMM) FLAME~\cite{li2017learning} into the structure model of FOMM to model the face region. FLAME models not only the face shape but also the expressional motion of eyelids, mouth and jaw, which provides strong prior knowledge of the face structure. This knowledge further helps accurately distinguish the occluded area in the generated images. We further exploit the successful inpainting technique, contextual attention module~\cite{yu2018generative}, to recover the appearance feature at the perceived occluded area that is highly relevant to other unoccluded feature patches. In addition, we propose a novel Geometrically-Adaptive (DE)normalization (GADE) layer to integrate the 3DMM geometric embeddings with the recovered appearance, which is shown to further enhance the generation quality in terms of facial details.

We conduct extensive experiments to compare our method with state-of-the-art methods~\cite{wang2019few, zakharov2019few, zakharov2020fast, siarohin2020first}. Both qualitative and quantitative results show that our approach outperforms previous works, especially in synthesizing faces in case of large pose differences, translational/expressional motions and occlusions.

In summary, our contributions include:
\begin{itemize}
  \item We combine the 2D affine motion model with 3DMM to elaborately model the scene structure in face images.
  \item We propose to exploit inpainting techniques to recover appearance feature at the perceived occluded area. To the best of our knowledge, this is the first combination of the two computer vision subjects of face animation and inpainting. 
  \item We propose a novel geometrically-adaptive denormalization layer to make full use of the 3D face geometry information for generating realistic facial details.
\end{itemize}

\section{Related Work}

\subsection{3D morphable model (3DMM)}
3D morphable model (3DMM) is a statistical model which transforms the shape and texture of a 3D face into a vector space representation~\cite{blanz1999morphable}. Traditional 3DMMs~\cite{blanz1999morphable, paysan20093d, cao2013facewarehouse, booth20163d, li2017learning} are low-dimensional linear models generated through principal component analysis (PCA). A 3D face is represented by a linear combination of those orthogonal bases with largest eigenvalues. The state-of-the-art linear model, FLAME (Faces Learned with an Articulated Model and Expressions)~\cite{li2017learning} is learned from about 3800 3D face scans. Apart from shape and expression, FLAME also models four articulated joints (jaw, neck, eyeballs) and pose-dependent corrective blendshapes, which enables more flexible and expressive full head modeling. In this paper, we adopt FLAME as a decoder to model 3D faces and use a convolutional neural network as an encoder to estimate 3DMM parameters from face images.

\subsection{Face animation}
Recent works on face animation can be divided into three categories: (1) motion-based methods, (2) landmark-based methods and (3) 3D-based methods.

Motion-based methods explicitly models a relative motion field from source to driving during the generation process. Early motion-based methods, e.g. X2Face~\cite{wiles2018x2face}, directly estimate a dense motion field by deep neural networks. Then the source face embedding is warped by the dense motion field to generate reenactment result. Monkey-Net~\cite{siarohin2019animating} and its follow-up work, First Order Motion Model (FOMM)~\cite{siarohin2020first} uses abstract self-supervised 2D keypoints to estimate multiple local sparse motions and then combines them into a dense motion field to warp the source image. One-Shot Talking Heads~\cite{wang2020one} extends the 2D keypoints to the 3D space and estimates a dense motion field in 3D, which enables more expressive and flexible motion modeling. \cite{siarohin2021motion} replaces the keypoints with regions and models in-plane rotation and scaling via principal component analysis (PCA). However, it still lack the capability to model out-of-plane rotations and expressional motions. Among all the motion-based works, only FOMM~\cite{siarohin2020first} and \cite{siarohin2021motion} models the occlusion using an occlusion map learned in a self-supervised way. In this paper, two categories of occluded area are perceived separately and recovered with the help of the 3D geometry embedding and the widely used inpainting technique.

Landmark-based methods refer to those works that utilize facial landmarks as conditions to synthesis animated source with generative models. Few-shot Talking Heads~\cite{zakharov2019few} uses landmarks as the conditional input and injects source appearance features to the generator through an adaptive instance normalization layer (AdaIn)~\cite{huang2017arbitrary}. Fast Bi-layer~\cite{zakharov2020fast} adopts spatially-adaptive denormalization (SPADE) layers~\cite{park2019semantic} to build a neural rendering system based on the landmark skeleton. Few-shot Vid2Vid~\cite{wang2019few} uses SPADE residual blocks to inject landmark features into the generator, where appearance features are adapted to driving landmarks. Even landmarks are possible to represent face poses and relative motions, it still lacks the ability to preserve identity information and to handle occlusion.

3D-based methods take advantages of the geometric prior of 3D faces, which are reconstructed by fitting 3DMM parameters to the source and driving frames. The traditional 3D method Face2face~\cite{thies2016face2face} transfers driving expressions to the source through deformation transfer~\cite{sumner2004deformation} and the reenacted 3D face is rendered to synthesize videos. In the deep learning era, 3D-based methods are combined with convolutional neural networks to generate more photo-realistic videos. HeadGAN~\cite{doukas2020headgan} uses the rendered 3D faces of the driving frames as conditional inputs to build a neural rendering generator, where both deformed source appearance features and driving audio features are injected to generate realistic animation videos. StyleRig~\cite{tewari2020stylerig} and GIF~\cite{GIF2020} manipulate face images through a pre-trained StyleGAN conditioned on 3DMM parameters. Basically, these methods can generate realistic images, but are intrinsically disadvantageous in spatial and temporal consistency on videos.

In this paper, we propose a method that combines the advantage of motion-based and 3D-based methods which can generate photo-realistic and identity-preserving reenactments even in case of large pose deviation and severe occlusion.

\section{Preliminaries}
\subsection{3D face prior}
We use FLAME~\cite{li2017learning} as the 3D face prior model in our framework. As a statistical 3D morphable model, FLAME acts as a decoder, which takes shape (or identity) $\mathbf{\alpha} \in \mathbb{R}^{\left | \mathbf{\alpha}  \right | } $, expression $\mathbf{\beta} \in \mathbb{R}^{\left | \mathbf{\beta}  \right | } $ and pose $ \mathbf{\theta } \in \mathbb{R}^{\mathrm{3k+3} } $ ($k$ joint rotations and one global rotation) parameters as inputs and outputs a 3D head mesh with $v=5023$ vertices and $f=9976$ faces. The model is mathematically defined as follows,
\begin{equation}
M(\mathbf{\alpha},\mathbf{\beta},\mathbf{\theta}) = W(T_{P}(\mathbf{\alpha}, \mathbf{\beta}, \mathbf{\theta}), \mathbf{J}(\mathbf{\alpha}), \mathbf{w}),
\end{equation}
where $W(\cdot)$ is the blend skinning function, which takes the offseted template mesh $T_{P} \in \mathbb{R}^{v \times 3}$, joints position $\mathbf{J}(\mathbf{\alpha}) \in \mathbb{R}^{3k} $ and blendweights $\mathbf{w} \in \mathbb{R}^{k\times v} $ as inputs.

\subsection{Differentiable rendering}
We adopt the differentiable renderer in PyTorch3D \cite{ravi2020pytorch3d} to render the reconstructed 3D face. The differentiable renderer in our case rasterizes 3D vertex attributes into the 2D image space. The rendering function $\mathcal{R}$ can be mathematically formulated as
\begin{equation}
    I = \mathcal{R}(M(\mathbf{\alpha},\mathbf{\beta},\mathbf{\theta}), \mathbf{c}, A),
    \label{rendering}
\end{equation}
where $I \in \mathbb{R}^{H\times W \times a}$ is the rendered 2D image. $\mathbf{c} = (s, \mathbf{t})$ represents the scale and translation parameters of the weak perspective camera model that we adopted in this paper. $A\in \mathbb{R}^{v\times a}$ represents 3D vertex attributes to fill on the image and $a$ is the attribute dimension. In the following of the paper, we use attributes of RGB texture, vertex normal and vertex offset (motion) between source and driving faces. We only keep the face region of the rendered image through a face mask in the UV space (eyes and mouth regions are not included). With the differentiable renderer, the whole framework can be trained end-to-end.

\begin{figure*}[h]
  \centering
  \includegraphics[width=0.95\linewidth]{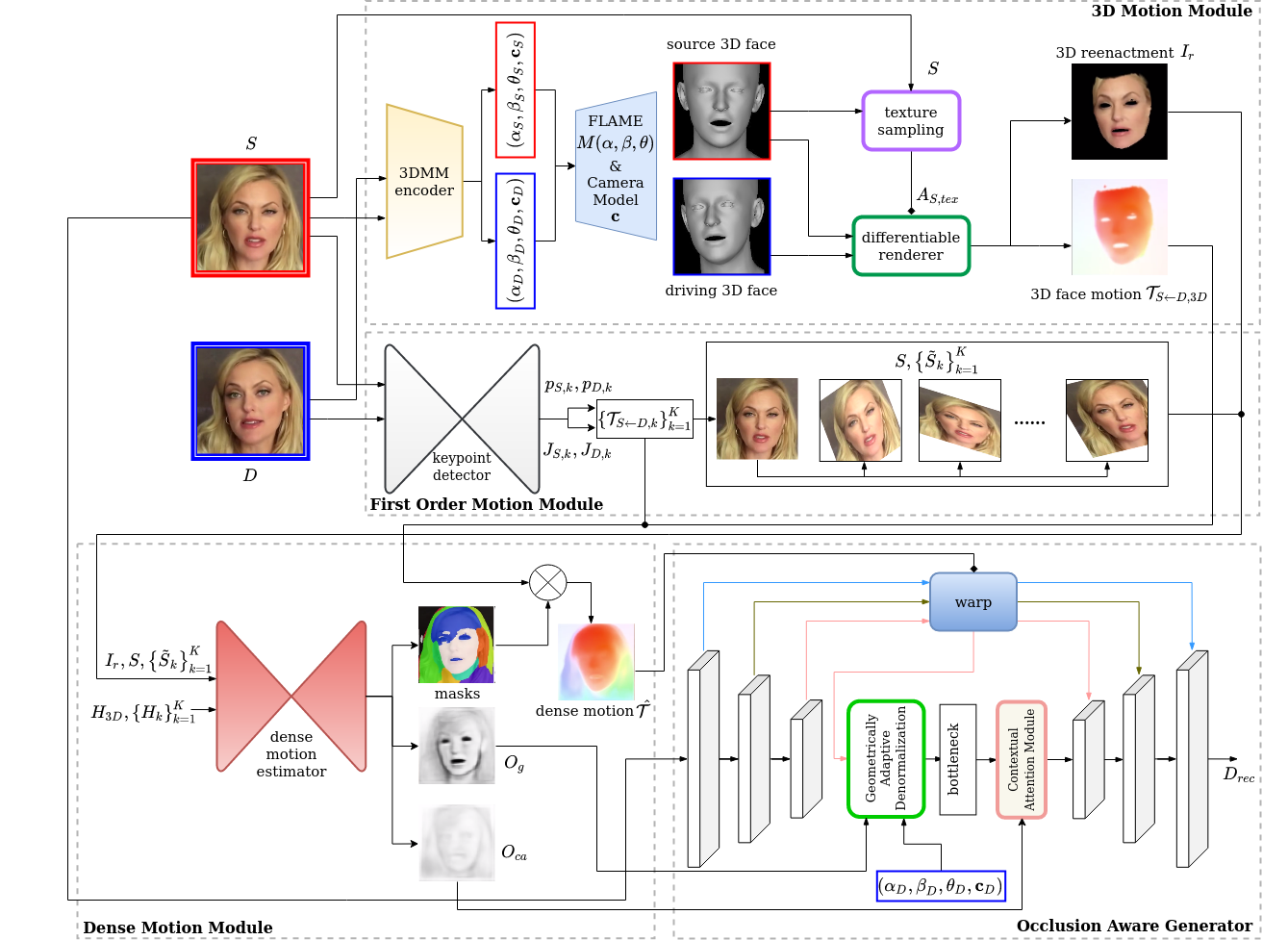}
  \caption{Overview of our framework. Our method takes a source image and a driving image as inputs. The 3D motion module reconstructs source and driving 3D faces and models 3D face motion using a differentiable renderer. The first order motion module dedicates to model the rest motions in terms of 2D local affine transformations. The dense motion module integrates both 3D and 2D motions and perceives where occlusion might happen. Finally, the U-Net shaped occlusion aware generator aligns the source image to the driving image.} 
  \label{overview}
\end{figure*}

\section{Method}

Our method takes a source face image $S\in \mathbb{R}^{H\times W \times 3}$ and a driving face image $D\in \mathbb{R}^{H\times W \times 3}$ as inputs and outputs the reenacted image $D_{rec}$ that mimics the face pose and expression of the driving face. As shown in Figure~\ref{overview}, our model contains four modules: (1) 3D motion module, (2) first order motion module, (3) dense motion module, (4) occlusion aware generator. In (1), we estimate the 3DMM parameters $\Phi = (\mathbf{\alpha},\mathbf{\beta},\mathbf{\theta}, \mathbf{c})$ of the source and the driving images and render a 3D face motion field $\mathcal{T}_{S\gets D, 3D}$ from driving to source. In (2), we predict $K$ abstract keypoints and Jacobians from source and driving images respectively and approximate $K$ local motion fields $\left \{ \mathcal{T}_{S\gets D, k}\right \} _{k=1}^{K}$ from driving to source. (3) integrates all the estimated local motions into a dense motion field $\hat{\mathcal{T}}$ and estimates two occlusion maps, $O_g$ and $O_{ca}$. (4) takes $S$, $\hat{\mathcal{T}}$, $O_g$, $O_{ca}$ and the driving 3DMM parameters $\Phi_D$ as inputs and generates the reenacted image. In the following subsections, we will discuss the four modules in detail.



\subsection{3D motion module} \label{3D motion module}
The 3D motion module contains a 3DMM encoder, a FLAME decoder and a differentiable renderer. The 3DMM encoder is pre-trained to predict the 3DMM parameters $\Phi$ given face images. 

Given $\Phi_S$ and $\Phi_D$, we obtain the 3D face vertices in the source and driving images from the FLAME decoder and then compute RGB texture, normal and motion attributes of these vertices. We interpret the source image $S(*)$ and the camera projection function $\mathbf{Proj}(*, \mathbf{c})$ as functions supporting arrayed inputs and outputs for convenience. The texture attributes $A_{S, tex}\in \mathbb{R}^{v \times 3}$ of the source image are obtained as
\begin{equation}
    A_{S, tex} = S(\mathbf{Proj}(M(\mathbf{\alpha}_{S},\mathbf{\beta}_{S},\mathbf{\theta}_{S}), \mathbf{c}_{S})),
\end{equation}
where bi-linear interpolation is used to obtain colors from $S$. The normal attributes $A_{S, n} \in \mathbb{R}^{v \times 3}$ and $A_{D, n} \in \mathbb{R}^{v \times 3}$ of both images (Omitted in the Figure \ref{overview}) are computed from the neighbor faces of each vertex. The vertex motion attributes $A_{m} \in \mathbb{R}^{v \times 2}$ are computed as
\begin{small}
\begin{equation}
    A_{m} = \mathbf{Proj}(M(\mathbf{\alpha}_{S},\mathbf{\beta}_{S},\mathbf{\theta}_{S}), \mathbf{c}_{S}) - \mathbf{Proj}(M(\mathbf{\alpha}_{D},\mathbf{\beta}_{D},\mathbf{\theta}_{D}), \mathbf{c}_{D}).
\end{equation}
\end{small}
Then by using Equation~\ref{rendering}, we rasterize the following maps:
\begin{gather}
    I_{r} = \mathcal{R}(M(\mathbf{\alpha}_D,\mathbf{\beta}_D,\mathbf{\theta}_D), \mathbf{c}_D, A_{S, tex}),\\
    I_{n, S} = \mathcal{R}(M(\mathbf{\alpha}_S,\mathbf{\beta}_S,\mathbf{\theta}_S), \mathbf{c}_S, A_{S, n}),\\
    I_{n, D} = \mathcal{R}(M(\mathbf{\alpha}_S,\mathbf{\beta}_D,\mathbf{\theta}_D), \mathbf{c}_D, A_{D, n}),\\
    \mathcal{T}_{S\gets D, 3D} = \mathcal{R}(M(\mathbf{\alpha}_{D},\mathbf{\beta}_{D},\mathbf{\theta}_{D}), \mathbf{c}_{D}, A_{m}). \label{3D motion}
\end{gather}
3D reenactment $I_{r}\in \mathbb{R}^{H\times W \times 3}$ refers to the rendered driving 3D face with texture transferred from source 3D face. $I_{n, S}\in \mathbb{R}^{H\times W \times 3}$ and $I_{n, D}\in \mathbb{R}^{H\times W \times 3}$ denote the normal maps of source and driving images, respectively. $\mathcal{T}_{S\gets D, 3D}\in \mathbb{R}^{H\times W \times 2}$ denotes the sparse motion field of the face area from the driving image to the source image. Exemplar visualization of $I_r$ and $\mathcal{T}_{S\gets D, 3D}$ is shown in Figure~\ref{overview}.

\subsection{First order motion module} \label{First order motion module}
We adopt the same sparse motion estimator as proposed in FOMM \cite{siarohin2020first}, where an Hourglass Network \cite{newell2016stacked} is used. The sparse motion estimator estimates $K=10$ keypoints $\left \{ p_{S, k}, p_{D, k} \in \mathbb{R}^{2} \right \}_{k=1}^{K}$ and Jacobians $\left \{ J_{S, k}, J_{D, k} \in \mathbb{R}^{2 \times 2} \right \}_{k=1}^{K}$ from the source and driving images respectively. The keypoints are represented in terms of heatmaps and Jacobians are calculated by average pooling of the four output channels. The local affine transformations are approximated by the first order Taylor expansion,
\begin{equation} \label{2D affine}
    \mathcal{T}_{S\gets D, k}(z) \approx p_{S,k} + J_{S,k}^{} J_{D,k}^{-1}(z - p_{D,k}),
\end{equation}
where $z \in \mathbb{R}^{2} $ is an arbitrary point on the driving frame. $K$ sparse affine motion fields $\left \{ \mathcal{T}_{S\gets D, k}(z)\right \} _{k=1}^{K}$ are used to warp the source image separately. Warping is simply implemented using a differentiable bi-linear sampling operator \cite{jaderberg2015spatial}. Besides the $K$ sparse motions, an identity motion field is appended to model the static background area. 

\subsection{Dense motion module} \label{densemotion}
The dense motion estimator is an encoder-decoder network with similar design to~\cite{siarohin2020first}. It takes deformed source images and a set of heatmaps as inputs, which are downsampled to $1/4$ of the original image resolution. The deformed source images include $I_r$, $S$, and $\left \{ \tilde{S}_k \right \} _{k=1}^{K}$. $\tilde{S}_k$ is obtained by deforming $S$ using $\mathcal{T}_{S\gets D, k}$. $I_r$ corresponds to the face motion computed from the 3D face model. $S$ corresponds to staticity of the background. $\tilde{S}_k$ corresponds to the 2D affine motion. The heatmaps indicate where motion might happen and also have $K + 2$ instances. $K$ heatmaps corresponding to the affine motion are formulated in the same way as FOMM, which are calculated by the difference of the heatmaps centered in $p_{D,k}$ and $p_{S,k}$.
\begin{small}
\begin{equation}
    H_{k}(z) = exp(\frac{-(p_{D,k} - z)^2}{2\sigma} ) - exp(\frac{-(p_{S,k} - z)^2}{2\sigma} ),
\end{equation}
\end{small}
where $\sigma = 0.01$. For the static background, the heatmap is represented by a zero map. For the heatmap of 3D face motion, we make use of the $z$ channel of the normal maps $I_{n,S}$ and $I_{n,D}$. 
The $z$ component indicates whether the 3D face is visible in the camera frame, which helps estimate not only the 3D motions but also the occlusion maps.
\begin{equation}
    H_{3D}(z) = I_{n,D}^{Z}(z) - I_{n,S}^{Z}(z).
\end{equation}

The dense motion module estimates $K + 2$ soft masks $M_{3D}$, $M_b$, and $\left \{  M_k \right \}_{k=1}^{K}$ (as shown in the bottom-left block of Figure \ref{overview}). $M_{3D}$ corresponds to the rigid 3D face (in white). $M_b$ corresponds to the static background (in black) and $\left \{  M_k \right \}_{k=1}^{K}$ correspond to 2D affine motion areas (in other colors). All the masks are softmaxed to guarantee their pixel-wise sum to be 1. Sparse motion fields are fused into a dense motion field through masked averaging:
\begin{small}
\begin{equation}
    \hat{\mathcal{T}}(z) = M_b z + M_{3D} \mathcal{T}_{S\gets D, 3D}(z) + \sum_{k=1}^{K} M_k \mathcal{T}_{S\gets D, k}(z).
\end{equation}
\end{small}
Besides, the dense motion module also predicts two occlusion maps $O_g, O_{ca} \in \left [ 0, 1 \right ]^{\frac{H}{4} \times \frac{W}{4}}$ indicating the invisible region in the source image but exposed in the generated image, which are used separately in the two elaborately designed modules in the occlusion aware generator to generate geometrically realistic face appearance and perceive areas need to be inpainted.

\subsection{Occlusion aware generator} \label{o-a generator}

The occlusion aware generator takes $S$, $\hat{\mathcal{T}}$, $O_g$, $O_{ca}$ and the driving 3DMM parameters $\Phi_D$ as inputs and generates the reenacted image $D_{rec}$. We adopt a U-Net \cite{ronneberger2015u} shaped network, which contains an encoder, a geometrically-adaptive denormalization layer (GADE), a bottleneck block, a contextual attention module and a decoder. The encoder takes the $S$ as input and outputs source feature maps of different scales. Then those feature maps are warped by $\hat{\mathcal{T}}$. The highest level warped features are ported to the GADE layer, the bottleneck block and the contextual attention module to further recover the occluded area. The GADE layer and the contextual attention module are guided by $O_g$ and $O_{ca}$ separately to avoid potential conflicts.  Finally, the recovered feature map is fed to the decoder. Lower level warped features are also concatenated to their corresponding intermediate features in the decoder in a U-Net fashion. Figure~\ref{overview} shows this procedure in the bottom-right block.


\subsubsection{Geometrically-adaptive denormalization}
\begin{figure}
  \centering
  \includegraphics[width=0.95\linewidth]{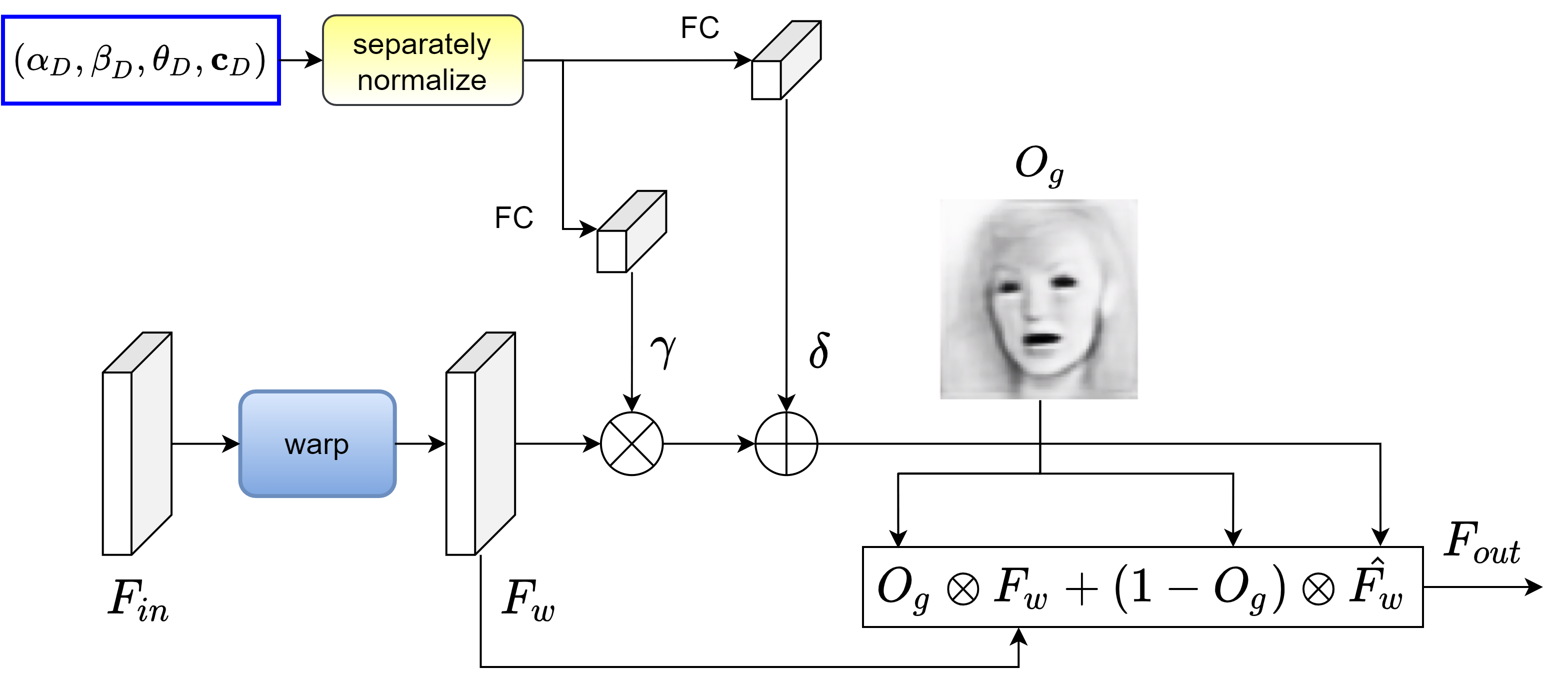}
  \caption{Illustration of the Geometrically-Adaptive Denormalization Layer}
  \label{GADE}
\end{figure}

The geometrically-adaptive denormalization (GADE) layer takes normalized features from the encoder as input and denormalizes the features on the condition of the driving 3D face geometry. Figure \ref{GADE} depicts the detail of GADE. We first normalize $\alpha, \beta, \theta, \mathbf{c}$, separately for numerical stability. $\alpha, \beta, s$ are divided by their average norms calculated from the training set and $\theta, \mathbf{t}$ are unchanged. Then we adopt two fully connected layers to convert these 3DMM parameters to a scale vector $\mathbf{\gamma}$ and a shift vector $\mathbf{\delta}$. 
The warped source feature $F_w$ is denormalized as
\begin{equation}
    \hat{F_w} = \mathbf{\gamma} \otimes F_w + \mathbf{\delta},
\end{equation}
where $\otimes$ denotes the broadcastable element-wise multiplication. The occlusion map $O_g$ is used to adaptively select between $F_w$ and $\hat{F_w}$,
\begin{equation}
    F_{out} = O_g \otimes F_w + (1-O_g) \otimes \hat{F_w}.
\end{equation}
In the original FOMM, $F_{out}$ only contains the first term and $F_w$ is simply diminished where occlusion occurs, i.e. $O_g$ is small. In our approach, we combine $\hat{F_w}$ in this case to exploit information from the driving image to recover the occlusion area. The injected face geometry code depicts 3D face shape, pose and expression of the driving image, which makes it easier to generate realistic features and to follow the driving face pose accurately. Note that we use $\Phi_D$ instead of its precedent features from the 3DMM encoder. This is because the precedent features might contain appearance information of the driving image and might cause the reenacted image to collapse to reconstruct the driving image without using the source image features. Denormalized features from GADE are fed to a bottleneck block that contains several residual convolution blocks, then used as inputs to the contextual attention module. 

\subsubsection{Contextual attention module}
\begin{figure}
  \centering
  \includegraphics[width=0.95\linewidth]{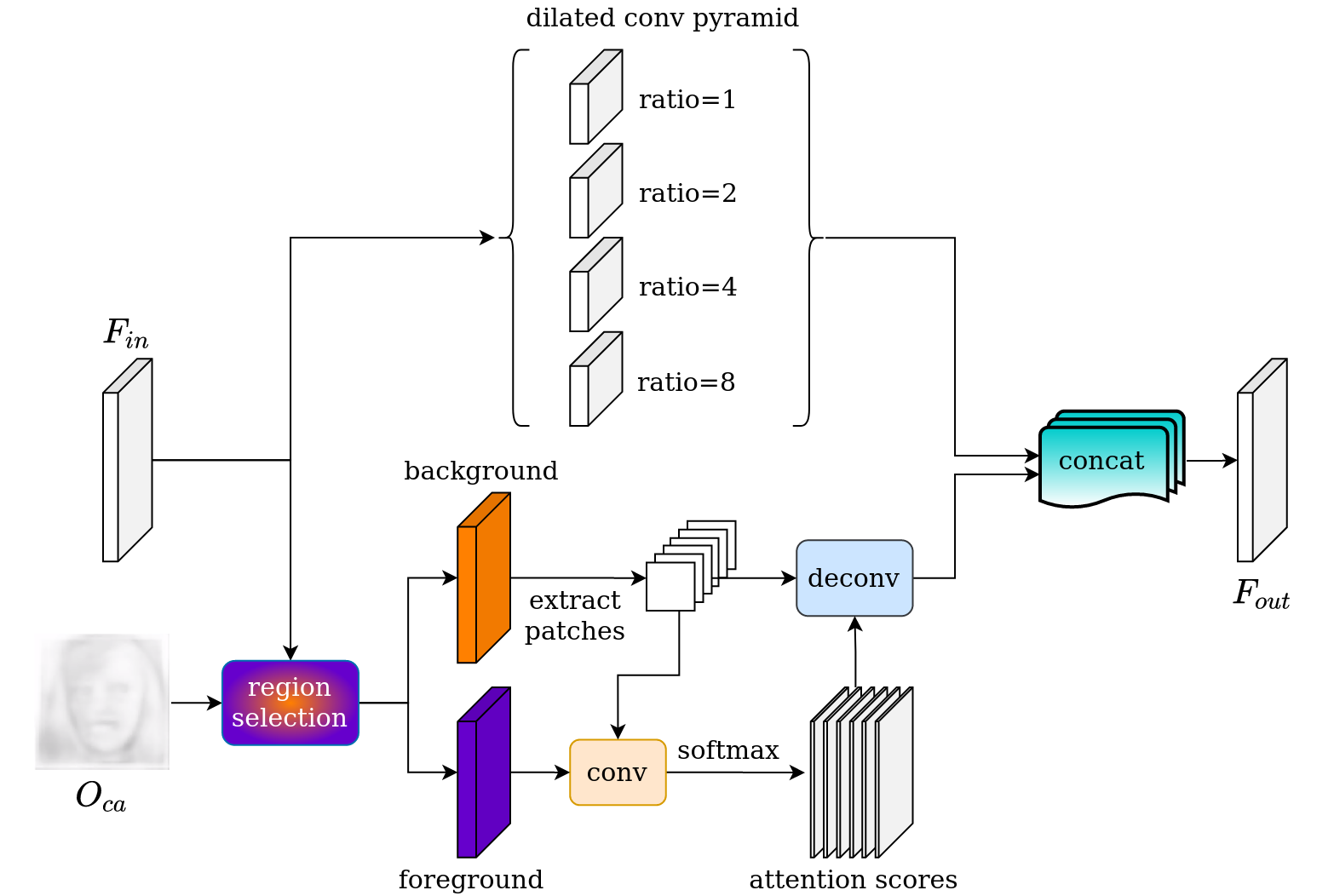}
  \caption{Illustration of the Contextual Attention Module}
  \label{CA}
\end{figure}

The original Contextual Attention Module (CAM) proposed in \cite{yu2018generative} was used for image inpainting, which takes a feature map and a hard binary mask indicating the regions to be inpainted as input. In our case, we adopt this module to realistically recover the perceived occluded areas indicated by the soft occlusion map $O_{ca}$ from the dense motion module. Note that we do not use face inpainting methods such as \cite{li2020learning} since we need to recover both the foreground and the background areas. The contextual attention module contains two parallel branches: (1) contextual attention layer, (2) dilated convolution block. The contextual attention layer is a differentiable layer. Patches are extracted from output features from the bottleneck block and each patch is used as a kernel to compute a matching score with the occluded area through convolution. Then softmax function is applied to the convoluted features to get attention scores. Finally occluded areas are reconstructed through deconvolution on the scores with the related feature patches. The parallel branch, dilated convolution block contains pyramid dilated convolution layers of dilation ratios $1, 2, 4, 8$. It not only increases the reception field but also enables the network to hallucinate according to neighbor regions. The output from contextual attention layer and the dilated convolution block is then concatenated and fed to the decoder of the generator to obtain the final reenactment $D_{rec}$. Figure \ref{CA} depicts the detail of the contextual attention module.

\subsection{Training}
The 3DMM encoder is pre-trained before we train the whole model end-to-end. We project 68 3D facial landmarks on the surface of the reconstructed 3D face $l_{i} \in \mathbb{R}^{3}$ into 2D image space and use the 2D landmark prediction $k_{i} \in \mathbb{R}^{3}$ from FAN \cite{bulat2017far} to formulate the $L1$ landmark re-projection loss. Besides, we use $L2$ parameter loss to regulate the 3DMM shape parameters $\mathbf{\alpha}$ and expression parameters $\mathbf{\beta}$ to obtain reasonable reconstruction results. 

Then, our framework is trained in a self-supervised strategy. We perform self-reenactment, where a pair of source and driving images are sampled from each video. The driving images are used as the ground truth for supervision. We adopt loss functions of two main objectives: (1) to improve the faithfulness and quality of output images; (2) to regulate the 3D motion module and the first order motion module to predict reasonable motion fields. For the first objective, we follow the same loss functions in FOMM, including perceptual loss \cite{johnson2016perceptual} and GAN-related loss. To regulate 3D motion, we preserve the 3DMM loss function in the pre-train stage. To stabilize 2D sparse keypoints and jacobians, we adopt the equivariance constraints from FOMM. Please refer to the supplementary material for the detailed expressions of loss functions.

\begin{figure*}
  \centering
  \includegraphics[width=0.9\linewidth]{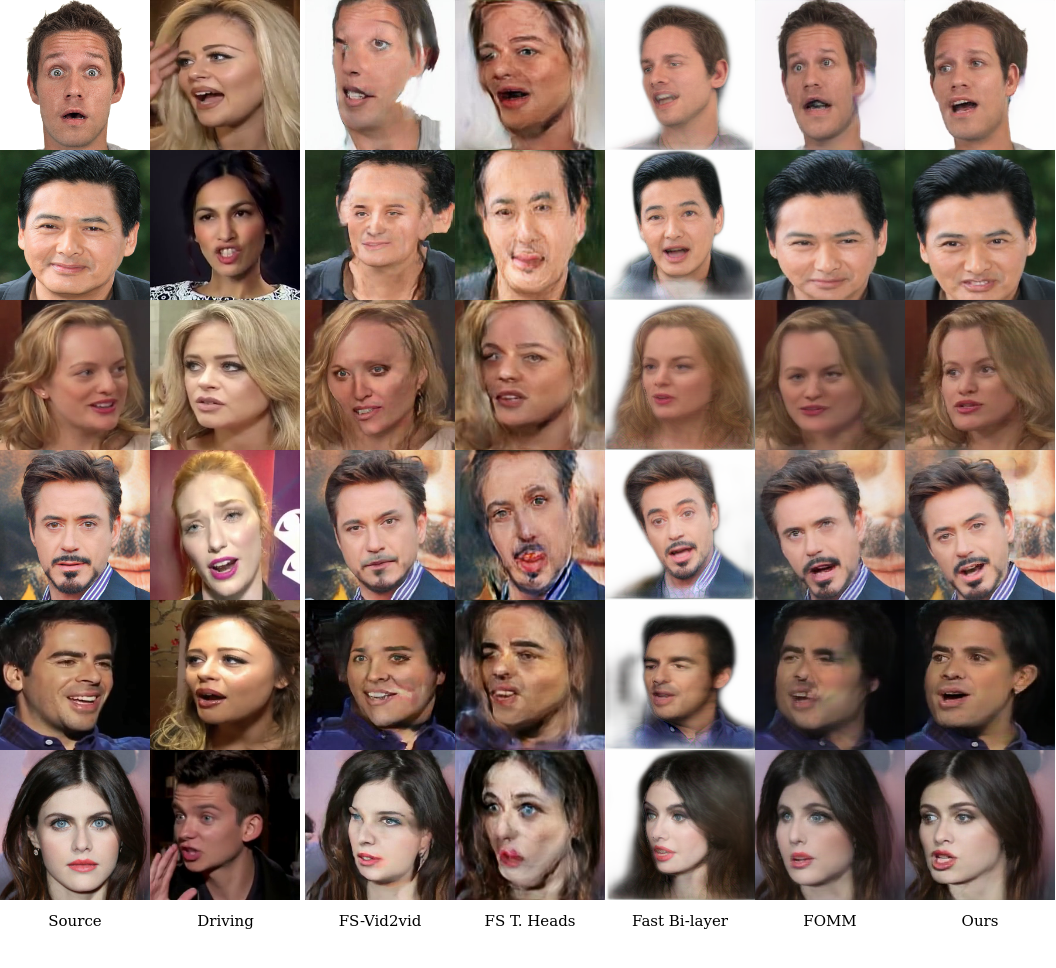}
  \caption{Qualitative comparison with SOTA methods on motion transfer.}
  \label{quali-compare}
\end{figure*}

\section{Experiments}
\subsection{Dataset}
We use Voxceleb1 \cite{Nagrani17} dataset to train and evaluate our approach. The Voxceleb1 dataset contains 22496 videos of 1251 celebrities extracted from YouTube videos. We adopt the same pre-processing strategy and train-test split as that of the FOMM. Videos of size $256 \times 256$ are cropped from the original videos covering the heads. In total, we obtain 17941 videos for training and 506 videos for testing after cropping. Besides, we also extract images from the pre-processed videos and use FAN \cite{bulat2017far} to predict 68 2D facial landmarks of each image.

\subsection{Implementation details}
We first pre-train the 3DMM encoder using the images extracted from pre-processed videos for every 10 frame. We adopt the lightweight MobilenetV2 \cite{sandler2018mobilenetv2} as our 3DMM encoder and modify the output fully connected layer to output 156 dimensions of FLAME parameters and 3 dimensions of camera parameters. We train the 3DMM estimator for 50 epochs with batch size of 128. Adam optimizer is used with learning rate $2e-4$ and $\beta_{1} = 0.9, \beta_{2} = 0.999$

For the end-to-end model training, we perform self-reenactment on the pre-processed videos. We randomly sample a pair of source and driving images from each video and use the driving image as ground truth. The 3DMM encoder is jointly fine-tuned with the animation model during end-to-end training. We train the whole model for 50 epochs and for each epoch we repeat the video list for 150 times. We adopt synchronized Batch Normalization to further improve the performance. Adam optimizer is used with learning rate $2e-4$ and $\beta_{1} = 0.5, \beta_{2} = 0.9$ for all network modules. We use 8 24GB NVIDIA P40 GPUs to train our model.

\subsection{Evaluation Metrics}
To quantitatively evaluate our approach and the previous state-of-the-art methods, we perform video reconstruction experiments, where the first frame of each test video is used as the source image and the rest frames are used as driving images. The following metrics are computed:

(1) \textbf{L1 distance}: Averaged L1 distance between the reconstructed images and the ground truth images.

(2) \textbf{Average Keypoint Distance (AKD)}: We extract facial landmarks from the reconstructed and ground truth images using FAN \cite{bulat2017far} and compute the average distance between the two set of landmarks. AKD measures how accurately the generated face video follows the face poses of the driving video.

(3) \textbf{Average Euclidean Identity Distance (AEID)}: We adopt the widely used face recognition network, ArcFace~\cite{deng2019arcface} to extract identity features from both generated and ground truth images. Then we calculate the averaged euclidean distance between the two. AEID measures how much identity information is preserved.

(4) \textbf{Fréchet Inception Distance (FID)}: FID was proposed in \cite{DBLP:journals/corr/HeuselRUNKH17}, which measures how realistic the visual quality of the generated images is, compared with the real images. We adopt the implementation of \cite {Seitzer2020FID}, which uses a pre-trained InceptionV3 \cite{szegedy2016rethinking} network to extract features from generated and ground truth images and calculates distribution distance between the two sets of features.

(5) \textbf{Peak Signal-to-Noise Ratio (PSNR)}: PSNR measures the image quality between generated and real images in terms of the absolute mean squared error (MSE).

(6) \textbf{Structural Similarity Index (SSIM)}: SSIM measures the structural similarity between the reconstructed and real images, which is more robust to global illumination changes compared with PSNR.

The units of L1, AKD, and PSNR are RGB intensity, pixels, and dB, respectively. FID, AEID and SSIM have no units since they are defined perceptually.

\begin{table}
\footnotesize
  \caption{Quantitative comparison with SOTA methods on Voxceleb1 test set.}
  \label{quanti-compare}
  \setlength{\tabcolsep}{1.2mm}{
  \begin{tabular}{ccccccc}
     & L1$\downarrow$& AKD$\downarrow$& AEID$\downarrow$& FID$\downarrow$& PSNR$\uparrow$& SSIM$\uparrow$\\
    FS-Vid2vid \cite{wang2019few}& 0.0869 & 6.214& 0.3538& 18.490& 29.349& 0.559\\
    FS T. Heads \cite{zakharov2019few}& 0.1053 & 10.247& 0.3243& 27.629& 28.906& 0.507\\
    Fast Bi-layer \cite{zakharov2020fast}& 0.3775& 12.486& 0.3544& 109.187& 28.118& 0.319\\
    FOMM\cite{siarohin2020first} & 0.0448& 1.386& 0.1654& 14.219& 30.567& 0.764\\
    Ours& \textbf{0.0401}& \textbf{1.222}& \textbf{0.1532}& \textbf{8.546}& \textbf{31.079}& \textbf{0.783}\\
  \end{tabular}}
\end{table}
\begin{table}
\footnotesize
  \caption{Quantitative results of ablation study on Voxceleb1 test set.}
  \label{quanti-ablation-result}
  \setlength{\tabcolsep}{1.2mm}{
  \begin{tabular}{ccccccc}
    & L1$\downarrow$& AKD$\downarrow$& AEID$\downarrow$& FID$\downarrow$& PSNR$\uparrow$& SSIM$\uparrow$\\
    FOMM baseline& 0.0448& 1.386& 0.1654& 14.219& 30.567& 0.764\\
    FOMM+3D& 0.0437& 1.315& 0.1676& 13.894& 30.644& 0.776\\
    FOMM+3D+CAM& 0.0418& 1.306& 0.1642& 11.273& 30.965& 0.780\\
    Full model& \textbf{0.0401}& \textbf{1.222}& \textbf{0.1532}& \textbf{8.546}& \textbf{31.079}& \textbf{0.783}\\
  \end{tabular}}
\end{table}
\subsection{Comparison with state of the art}
We compare our approach with the state-of-the-art face animation methods FS-Vid2vid~\cite{wang2019few}, FS Talking Heads~\cite{zakharov2019few}, Fast Bi-layer~\cite{zakharov2020fast}, FOMM~\cite{siarohin2020first}, and One-Shot Talking Heads~\cite{wang2020one}. For FS-Vid2vid and Fast Bi-layer, we use the pre-trained models provided by the authors. But note that their training sets are much larger than ours (FS-Vid2vid: FaceForensics \cite{roessler2019faceforensicspp}, Fast Bi-layer: Voxceleb2 \cite{Chung18b}). For FS Talking Heads and FOMM, we train their models on the same training set as our method. For the latest proposed method One-Shot Talking Heads~\cite{wang2020one}, due to the unavailability of the official source code, we only provide qualitative comparisons using cases reported in their paper. 

\begin{figure}
  \centering
  \includegraphics[width=0.95\linewidth]{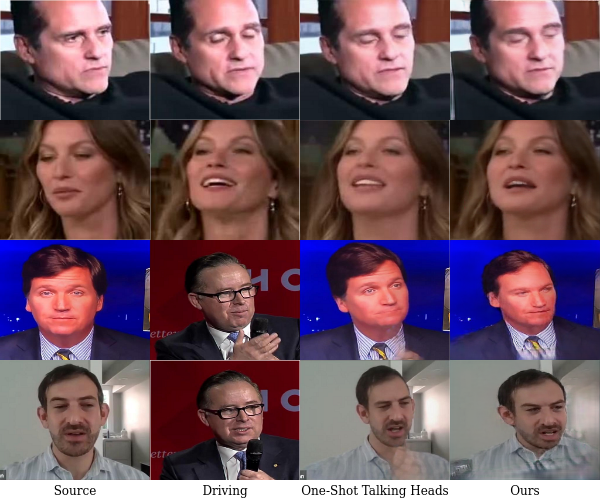}
  \caption{Qualitative comparison with \cite{wang2020one} on two successful (row 1-2) and two failed (row 3-4) cases.} 
  \label{one-shot compare}
\end{figure}

We provide quantitative evaluation results for video reconstruction on the test set of Voxceleb1 dataset, as listed in Table \ref{quanti-compare}. Our method performs best in all metrics. To visually illustrate the improvement besides the above abstract metrics, we provide qualitative visualizations on motion transfer, where source and driving faces are of different identities, as shown in Figure \ref{quali-compare}. We intentionally choose image pairs with large pose differences, translational motions and challenging expressions to show our superiority. The qualitative comparison with One-Shot Talking Heads is shown in Figure \ref{one-shot compare}. We generate face images using a similar relative motion transfer strategy as in FOMM, please refer to the supplementary material for the details of the inference process of the motion transfer. Our method shows superiority in generating realistic, pose-preserving and identity preserving face animation results. Moreover, we improves the image quality for the occluded failure cases by \cite{wang2020one} with fewer parameters and training data.

\begin{figure}
  \centering
  \includegraphics[width=0.95\linewidth]{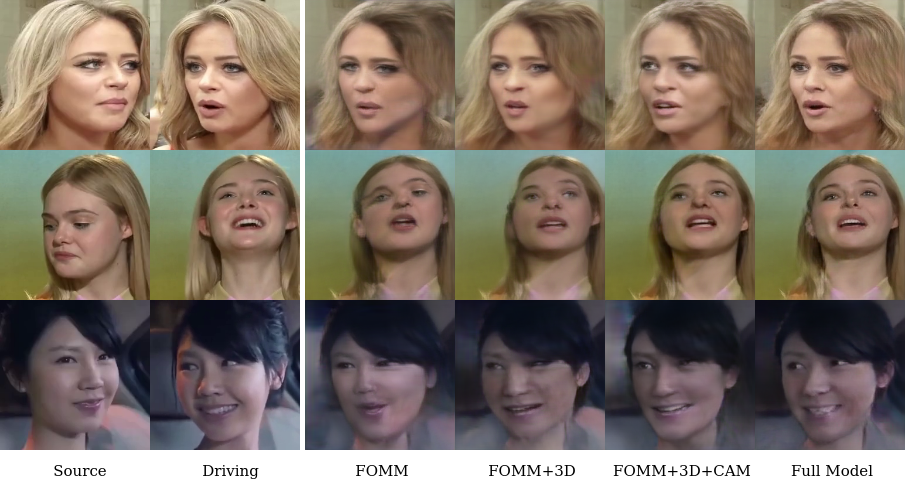}
  \caption{Qualitative results of ablation study on video reconstruction.}
  \label{quali-ablation}
\end{figure}

\subsection{Ablation study}
We conduct both quantitative and qualitative experiments on video reconstruction to show the contributions of the 3D motion module, the Contextual Attention Module (CAM) and the Geometrically-Adaptive Denormalization layer (GADE). Table \ref{quanti-ablation-result} shows the quantitative results on the test set of Voxceleb1. With the integration of 3D face prior knowledge, the AKD score decrease significantly, which means the face pose of the reenactment becomes more accurate. The additional contextual attention module further improves the FID scores with context-coherent inpainting. Finally, injection 3D face geometry information into the generation process benefits both image quality and face appearance of the generated image. This shows the effectiveness of 3D geometry embedding in learning the scene structures. We also provide qualitative results as shown in Figure \ref{quali-ablation}. Without 3D modelling, the faces might be improperly distorted. With the contextual attention module, the visual quality of the generated images improves. With the GADE layer, facial details and expressions become more realistic and self-occluded face area is properly recovered. The full model outperforms all the model variations in terms of visual sharpness of both foreground and background.

\section{Conclusions}
In this paper, we propose SAFA, a structure aware face animation method. It enhances the animation quality by exploiting the scene structure knowledge to model each component in the image. We show the effectiveness of combining 3DMM with the 2D affine motion model in modelling the foreground. Besides, the contextual attention module for image inpainting further helps generate realistic images by recovering the perceived occluded area. Moreover, the proposed geometrically-adaptive denormalization layer boosts the quality of face appearance by injection of 3D geometry information. Both quantitative and qualitative experiments show that our method outperforms the state-of-the-art methods.

{\small
\bibliographystyle{ieee_fullname}
\bibliography{egpaper_final}
}
\newpage

\begin{appendices}
\section{Details of loss function}

\textbf{Perceptual loss} We adopt a pre-trained VGG-19 network \cite{simonyan2014very} to extract features of $l$ layers from the driving image $D$ and the reconstructed image $D_{rec}$. Original images are downsampled $4$ times to extract a feature pyramid. Then $L1$ loss is applied to the two sets of feature maps.
\begin{small}
\begin{equation}
   \mathcal{L}_{p} = \sum_{i=1}^{4} \sum_{j=1}^{l}\left \| f_{VGG}^{i, j}(D) - f_{VGG}^{i, j}(D_{rec}) \right \|_{1}.
\end{equation}
\end{small}

\textbf{GAN loss} We adopt a multi-scale discriminator $\mathbf{D}_{m}$ as used in \cite{wang2018high}. The driving image and the reconstructed image of multiple resolutions are fed into the discriminator. We use the least square loss proposed in \cite{mao2017least} to enable high-quality image generation and stable training.
\begin{small}
\begin{gather}
\mathcal{L}_{G} = \sum_{i=1}^{s} \mathbb{E} \left [ (1 - \mathbf{D}_{m}^{i}(D_{rec}^{i}))^{2} \right ], \\
\mathcal{L}_{D} = \sum_{i=1}^{s}\mathbb{E} \left [(\mathbf{D}_{m}^{i}(D_{rec}^{i}))^{2} \right] + \mathbb{E} \left[(1 - \mathbf{D}_{m}^{i}(D^{i}))^{2} \right].
\end{gather}
\end{small}

\textbf{3DMM loss} The 3DMM loss contains a $L1$ landmark re-projection term and a $L2$ parameter regularization term.
\begin{equation}
\label{3DMM loss}
L_{3DMM} = \sum_{i=1}^{68}\left \| k_{i} - \mathbf{c}(l_{i}) \right \|_{1} + \lambda_{\alpha} \left \| \mathbf{\alpha} \right \| _{2}^{2} +  \lambda_{\beta} \left \| \mathbf{\beta} \right \| _{2}^{2},
\end{equation}
where $\lambda_{\alpha}$, $\lambda_{\beta}$ are weights of the regularization on shape and expression parameters. The loss is used in both pre-training and the end-to-end stage. 

\textbf{Equivariance constraints} We adopt this loss to ensure consistency of the sparse keypoints. Specifically, we randomly sample an affine transformation $\mathcal{T} (z)$ and apply it to the original driving image. If we transform the predicted keypoints from the deformed driving image back, the result should be sufficiently close to the keypoints estimated from the original driving image. We minimize the $L1$ loss between the two value, similar constrains can also be applied to the Jacobians.
\begin{small}
\begin{equation}
    \mathcal{L}_{e} = \sum_{k=1}^{K}\left \| p_{D,k} - T^{-1}(p_{D_T, k}) \right \|_{1}.
\end{equation}
\end{small}

In total, the final loss function is given by
\begin{equation}
    \mathcal{L} = \lambda_p \mathcal{L}_p +\lambda_G \mathcal{L}_G + \lambda_D \mathcal{L}_D + \lambda_{3DMM} \mathcal{L}_{3DMM} + \lambda_e \mathcal{L}_e,
\end{equation}
where $\lambda_{*}$ are weights that balance contributions of each loss term. The loss weights are set as $\lambda_p = 10, \lambda_G = 1, \lambda_D = 1, \lambda_{3DMM} = 1, \lambda_{\alpha} = 1e-2, \lambda_{\beta} = 8e-3, \lambda_e = 10$.

\begin{figure*}[h]
  \centering
  \includegraphics[width=0.96\linewidth]{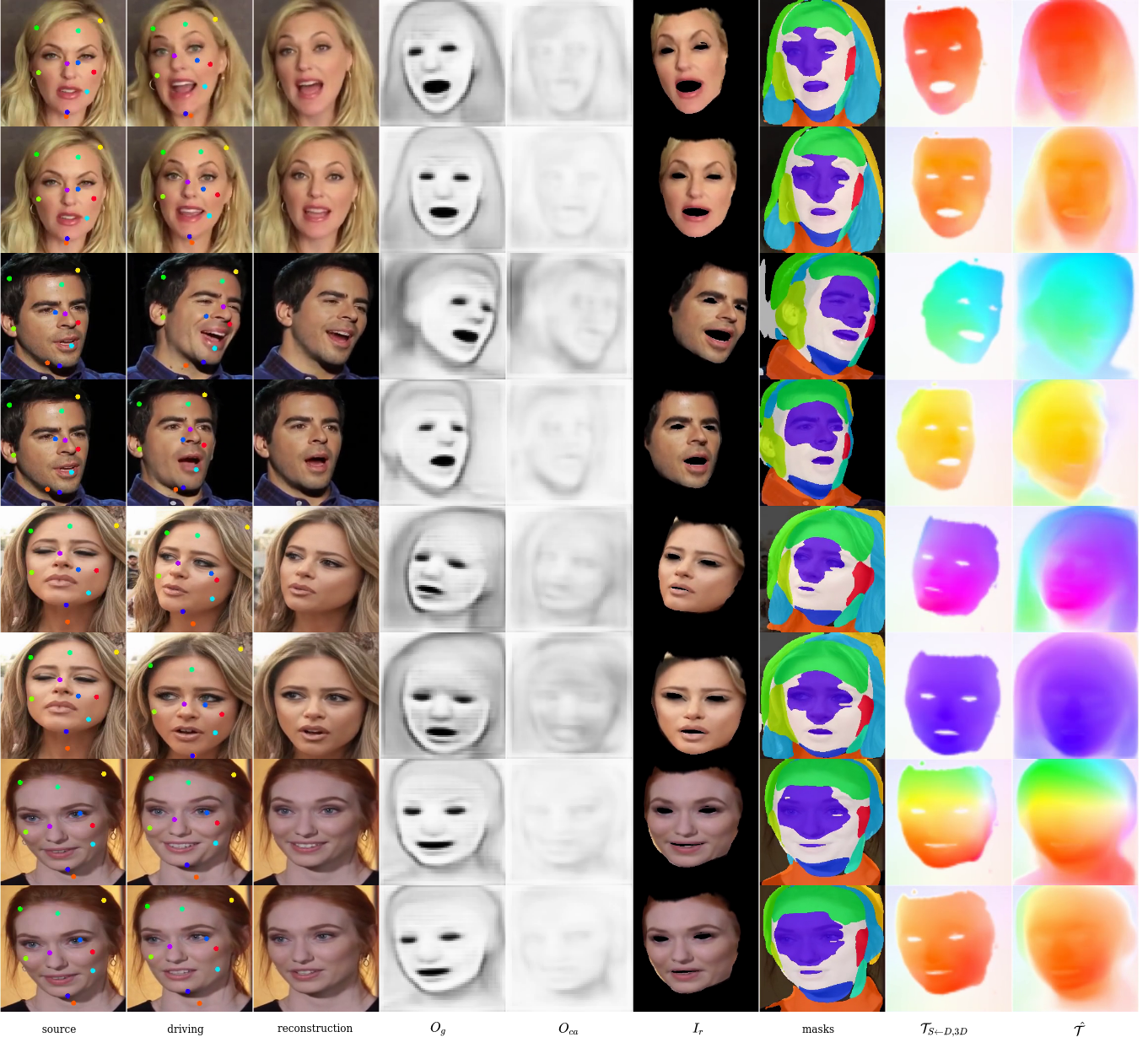}
  \caption{Visualization of intermediate components. From left to right are: source image with 2D keypoints, driving image with 2D keypoints, reconstructed image, occlusion map for GADE $O_g$, occlusion map for contextural attention module $O_{ca}$, 3D reenactment $I_r$, dense motion mask (the white part corresponds to $M_{3D}$, the black part corresponds to $M_b$ and others correspond to $\sum_{k=1}^{K} M_k$), 3D motion field $\mathcal{T}_{S\gets D, 3D}$ and dense motion field $\hat{\mathcal{T}}$} 
  \label{component_vis}
\end{figure*}

\section{Inference}
Our inference procedure is similar to that of FOMM. We first select a reference frame $D_r$ from the driving video, which has the nearest pose with respect to $S$. Then we impose the relative motion between the $D_r$ and each frame $D_t$ of the driving video on $S$ to generate the reenactment $S_{t}$. For the 2D affine motion, the relative motion $\mathcal{T}_{s \gets s_t, k}$ is represented as \footnote{$J_{S,k}$ is canceled out here.}
\begin{equation}
    \mathcal{T}_{S \gets S_t, k}(z) \approx p_{S,k} + J_{D_r,k}^{} J_{D_t,k}^{-1}(z - p_{S,k} + p_{D_r, k} - p_{D_t, k}).
\end{equation}
For the 3D face motion, we calculate the relative motion in terms of relative 3DMM parameters. The 3DMM parameters of the relatively reenacted image are
\begin{small}
\begin{equation}
    \Phi_{S_t} = (\mathbf{\alpha}_S, \mathbf{\beta}_S + \mathbf{\beta}_{D_t} - \mathbf{\beta}_{D_r}, \mathbf{\theta}_S + \mathbf{\theta}_{D_t} - \mathbf{\theta}_{D_r}, (s_S\frac{s_{D_t}}{s_{D_r}}, \mathbf{t}_S + \mathbf{t}_{D_t} - \mathbf{t}_{D_r})).
\end{equation}
\end{small}
Then the relative vertex motion attributes are
\begin{small}
\begin{equation}
    A_{m, S_t} = \mathbf{Proj}(M(\mathbf{\alpha}_{S},\mathbf{\beta}_{S},\mathbf{\theta}_{S}), \mathbf{c}_{S}) - \mathbf{Proj}(M(\mathbf{\alpha}_{S_t},\mathbf{\beta}_{S_t},\mathbf{\theta}_{S_t}), \mathbf{c}_{S_t}).
\end{equation}
\end{small}
Finally, the relative 3D face motion is rendered as 
\begin{equation}
    \mathcal{T}_{S \gets S_t, 3D}(z) = \mathcal{R}(M(\mathbf{\alpha}_{S_t},\mathbf{\beta}_{S_t},\mathbf{\theta}_{S_t}), \mathbf{c}_{S_t}, A_{m, S_t}).
\end{equation}


\section{Intermediate components visualization}
In Figure~\ref{component_vis}, we provide additional qualitative visualizations of the intermediate components produced by each module in our framework, including the 2D affine keypoints $\left \{ p_{S, k}, p_{D, k} \right \}_{k=1}^{K}$,  3D reenactment $I_r$, 3D face motion $\mathcal{T}_{S\gets D, 3D}$, motion masks $M_{3D}$, $M_b$, and $\left \{  M_k \right \}_{k=1}^{K}$, dense motion $\hat{\mathcal{T}}$, and two occlusion maps $O_g$ and $O_{ca}$ for the GADE layer and the contextual attention module respectively. The excellent accuracy of the 3DMM encoder can be demonstrated from the 3D reenactment $I_r$ (column 6).

\end{appendices}

\end{document}